# Clustering via torque balance with mass and distance

Jie Yang and Chin-Teng Lin*

University of Technology Sydney, Faculty of Engineering and Information Technology, Centre for Artificial Intelligence, 15 Broadway, Ultimo, New South Wales 2007, Australia

*Chin-Teng.Lin@uts.edu.au

**Abstract**

Grouping similar objects is a fundamental tool of scientific analysis, ubiquitous in disciplines from biology and chemistry to astronomy and pattern recognition. Inspired by the torque balance that exists in gravitational interactions when galaxies merge, we propose a novel clustering method based on two natural properties of the universe: mass and distance. The concept of torque describing the interactions of mass and distance forms the basis of the proposed parameter-free clustering algorithm, which harnesses torque balance to recognize any cluster, regardless of shape, size, or density. The gravitational interactions govern the merger process, while the concept of torque balance reveals partitions that do not conform to the natural order for removal. Experiments on benchmark data sets show the enormous versatility of the proposed algorithm.

**Methods**

"♫ One of these things is not like the others; one of these things just doesn't belong." Little did we know as children watching Sesame Street, how important the concepts conveyed by Ernie and Bert would be in our lives as adults, and particularly now in the Age of Technology. Grouping similar objects to derive insights from classes of things is a fundamental tool in the search for knowledge. It is used in virtually all natural and social sciences and plays a central role in biology, astronomy, psychology, medicine and chemistry. Like many disciplines, in data science, grouping objects is called clustering and, as one of the three broadest categories of machine learning algorithms, clustering in one form or another is the only method of learning from unlabelled data.

Yet, despite the importance and ubiquity of clustering and the plethora of existing clustering algorithms, they suffer from a variety of drawbacks and no universal solution has yet emerged [1]. To run through the list [2–4], hierarchical clustering has a high computational cost and a typical time complexity of at least $O(n^2)$; hence, these algorithms are not suitable for large-scale data sets. Further, stopping the clustering process requires some manually-determined condition, such as "Stop at k number of clusters".

Partition clustering demands that the number of clusters must either be known or estimated in advance, and these algorithms cannot detect non-convex clusters of varying size or density. Plus, they are highly sensitive to noise, outliers, and getting the initialization phase "right".

Density clustering requires a suite of thresholds to be set in advance – for example, the cutoff distance used to calculate the density of points, the number of points at which a cluster is deemed to be high density, and so on.

Model-based clustering generally relies on prior knowledge of many parameter settings, such as the distribution of each cluster, even though this information is often very difficult to acquire in practice.

Lastly, grid clustering also depends on many user-provided parameters, such as interval values to divide space and density thresholds, and the algorithms do not scale to high dimensional data sets.

A clustering algorithm is needed that can: recognize all kinds of clusters regardless of shape, size or density; is parameter-free; does not depend on a priori knowledge; has low computational overhead and reasonable time complexity; is robust to noise, outliers; does not need any initialization; and does not demand a manually-specified stopping condition. To achieves these goals, we propose a new clustering algorithm called torque clustering (TC), which is inspired by the torque balance that occurs during gravitational interactions in galaxy mergers [5,6]. The torque balance is founded on Newton's law of universal gravitation $F = \frac{Gm_1m_2}{r^2}$, which considers two natural properties in the universe: mass and distance.

These two properties exist in all areas of the natural sciences, which inspired us to wonder why not in clustering too? Setting $G$ as a constant aside for the moment, consider $m_1m_2$ and $r^2$, and imagine $m_1$ and $m_2$ as the number of samples in two data clusters and $r^2$ as the distance between them. Just as the individual galaxies in our universe are so obvious – being enormous numbers of densely-packed stars with vast empty space between them – cluster partitions can be determined in this very same way. Find groups of many samples separated by long distances, and there we have our clusters. However, finding "reasonable" clusters, at say a solar system level, requires a different law of physics and, here, we turn to torque balance. The concept of torque balance dictates that greater length is offset by less mass and vice versa. Therefore, if two clusters both contain many samples and the distance between them is also long, they cannot merge into a natural partition structure.

Loosely based on conventional hierarchical clustering structures [7], the proposed TC algorithm generates a hierarchical tree that reflects the natural structure of the data set. However, unlike traditional hierarchy-based algorithms, TC reaches higher accuracy with a significantly smaller number of mergers and does not require a manually determined condition, such as the number of clusters, to stop the clustering process. In addition, it has a low computational overhead, a reasonable time complexity, is robust to noise and outliers and does not require initialization.

Given a data set, each sample has a mass of one and, initially, each sample is its own cluster. Therefore, in the beginning, the mass of each cluster = 1. Then, the following rule is applied to form connections between clusters:

$$\zeta_i \rightarrow \zeta_i^N, if\ mass(\zeta_i) \leq mass(\zeta_i^N), \qquad (1)$$

where $\zeta_i$ denotes the *i*-th cluster, $\zeta_i^N$ denotes the 1-nearest cluster of $\zeta_i$, $mass(\zeta_i)$ is the number of samples the cluster contains and, since each sample has a mass of 1, it also represents the mass of $\zeta_i$. Similarly, $mass(\zeta_i^N)$ is the mass of $\zeta_i^N$. The symbol " $\rightarrow$ " denotes a connection $C_i$ from $\zeta_i$ to $\zeta_i^N$. The connection between two clusters represents the potential merge of these two clusters. By continuing this merger process according to Eq. (1), all clusters will eventually merge into one cluster and form a hierarchical tree. However, according to the concept of torque balance, if a connection has both relatively large mass and distance, it will be deemed "abnormal" and should be removed to reveal a more reasonable partition structure.

The abnormal connections can be identified by observing two intuitive properties of the connection $C_i$. One of the properties is the product of the mass of the two clusters it connects

$$M_i = mass(\zeta_i) \times mass(\zeta_i^N), \quad (2)$$

The other is the square of the distance between the two clusters it connects

$$D_i = d^2(\zeta_i, \zeta_i^N). \quad (3)$$

Plotting all the connections on a two-dimensional graph of the two properties, called the decision graph, will reveal that the mass and distance of the abnormal connections are abnormally larger than, and further away from, those of the normal connections.

Fig. 1 provides an example to illustrate the core idea in the proposed TC algorithm. Here, several clusters and their connections to their 1-nearest clusters are established based on the data distribution and the connection method in Eq. (1). Clusters G and H are connected by $C_5$, and Clusters A and B within Cluster G are connected by $C_1$. By calculating just two properties of each connection, $M_i$ and $D_i$, we can find the abnormal connections whose $M_i$ and $D_i$ are both relatively large. Connection $C_5$ meets this requirement, so we can remove it to obtain an appropriate cluster partition.

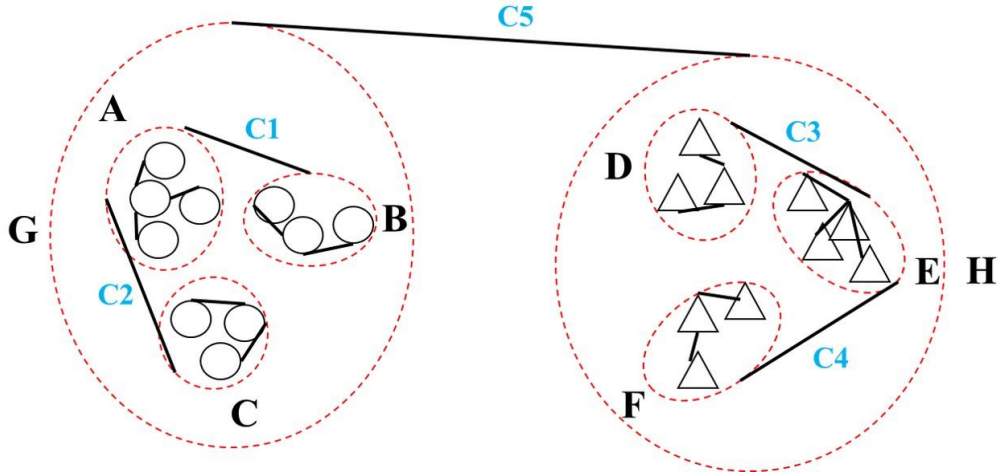

**Fig. 1. Torque balance as a 'threshold' for cluster partitions.** The red dotted lines delineate clusters A-H derived from Eq. (1) in this two-dimensional data distribution. The black lines $C_1$-$C_5$ indicate the connections from each cluster to its 1-nearest cluster with a length of $L_i$, where $L_5$ is the longest. Each cluster is at one end of a connection $C_i$, and contains several samples. For example, each of clusters A and E has four samples, each of clusters B, C, D, and F has three samples, and each of clusters G and H has 10 samples. Our goal is to find "abnormal" connections, defined as those with a distance $L_i$ (i.e., lever length) and a number of clusters (i.e., mass) that are both relatively large – concurrent conditions that conceptually disobey the rules of torque balance. Obviously, connection $C_5$ with 10 samples in each of its "galaxies" spread many light-years apart meets this goal and should be removed, leaving the "solar system"-sized clusters with only 3 or 4 samples as a more reasonable clustering scheme. This approach is consistent with human intuition as well as the natural laws of gravitational interactions.

To calculate the distance between a cluster and its 1-nearest cluster, we chose the single-linkage method [8], which measures the nearest distance from any member of one cluster to any member of the other cluster. But any preferred linkage method could be used, such as complete-linkage, average-linkage, and centroid-linkage [9,10]. With large-scale data, a fast approximate nearest neighbor method like k-d tree or locality-sensitive hashing may be a more appropriate choice since the distance computation approach used in these methods negates the need to actually know the distances between any two clusters [11,12]. Further, the computation costs would be low and the complexity could be kept to *O(nlog(n))*, as compared to the complexity of traditional hierarchical clustering algorithms, which is at least *O(n²)*. In this way, the proposed TC algorithm is highly scalable.

The details of how TC works is best explained through an example, which is set out step-by-step in Figure 2. The pseudocode is provided in Table S1 in the supplementary materials.

**Fig. 2. A step-by-step example of how TC works.** Consider a data set where, initially, each sample is its own cluster. Connections between clusters are then established according to Eq. (1), which results in a connected graph. By calculating the connected components in the graph, clusters begin to emerge in a way somewhat similar to the way galaxies and solar systems form [5]. Hence, the analogy which continues throughout this example. Galaxy formation begins with Fig. 2A. (Note that different colors correspond to different galaxies, and the number in the circle is the mass of each galaxy.)

### A

The first phase of the algorithm is to merge all galaxies into one universe. The distribution map of raw connections produced by Eq. (1) reveals three large and several small galaxies as indicated by the elliptical dotted lines.

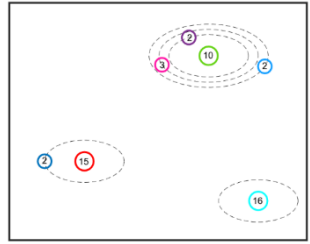

### B

Connections $C_1$-$C_4$ can then be added according to the 1-nearest relationship given by Eq. (1). Now, the two properties $M_i$ and $D_i$ can be calculated for each connection, as shown in Table 1a.

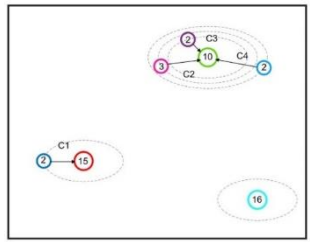

The mass of the galaxy at one end of a connection is $m_1$, the mass of the galaxy at other end is $m_2$ and the distance between the two is $d$. Therefore, each connection has two properties $m_1 \times m_2$ and $d^2$.

### C

The smaller galaxies are absorbed into the larger ones according to the 1-nearest relationship given in Table 1b and Eq. (1).

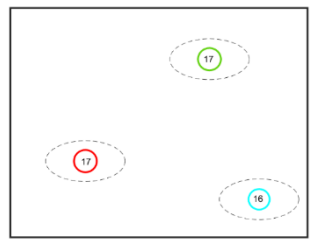

The mass of each new galaxies is equal to the sum of the masses of the subgalaxies it contains.

### D

Connections $C_5, C_6$ join the galaxies and $M_i$ and $D_i$ are calculated for each as per Fig. 2B, as shown in Table 1b.

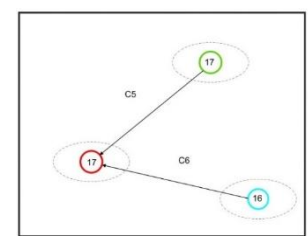

### E

Joining these together, we now have one big galaxy, and the merging process is complete, and a hierarchical tree is also established, as shown in Fig. 2H.

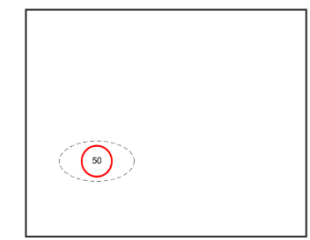

## F

Returning to Fig. 2D for a moment, it is easy to see from the decision graph in Fig. 2G that the relative maxima of the two properties that need to be removed are at $C_5, C_6$, which are identified as abnormal connections indicated by the red dotted line.

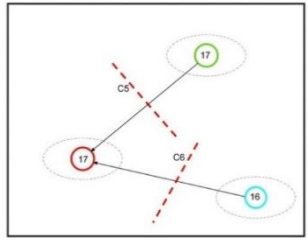

## G

Plotting all six connections on a two-dimensional graph of the properties, called the decision graph, indeed shows that $C_5, C_6$ are abnormally larger than, and further away from, $C_1 - C_4$.

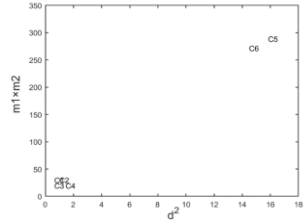

## H

Hence, connections $C_5, C_6$ are removed to arrive at the final partitioning scheme. This entire clustering process can be represented as a hierarchical tree, as the dendrogram to the right shows. The solid black arrows indicate the calculation of connected components.

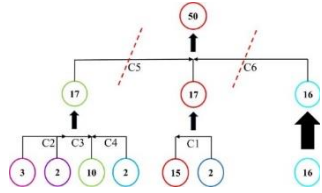

**Table 1a.** Properties of the galaxies in Figs. 2A and 2B.
Different colors correspond to different galaxies. "N" means that the galaxy does not connect to its 1-nearest galaxy, and "C" means it connects. $m_1$ and $m_2$ are the masses of a given galaxy and its 1-nearest galaxy, respectively. $d$ is the distance between a galaxy and its 1-nearest galaxy.

| Galaxy | Mass of the Galaxy | Nearest Galaxy | Mass of the Nearest Galaxy | Connect it or not? | $m_1 \times m_2$ | $d^2$ | Connect number |
|---|---|---|---|---|---|---|---|
| ○ | 15 | ○ | 2 | N | – | – | – |
| ○ | 2 | ○ | 15 | C | 30 | 0.64 | C1 |
| ○ | 10 | ○ | 2 | N | – | – | – |
| ○ | 3 | ○ | 10 | C | 30 | 1.00 | C2 |
| ○ | 2 | ○ | 10 | C | 20 | 0.64 | C3 |
| ○ | 2 | ○ | 10 | C | 20 | 1.44 | C4 |
| ○ | 16 | ○ | 15 | N | – | – | – |

**Table 1b.** Properties of the galaxies in Figs. 2C and 2D.

| Galaxy | Mass of the Galaxy | Nearest Galaxy | Mass of the Nearest Galaxy | Connect it or not? | $m_1 \times m_2$ | $d^2$ | Connect number |
|---|---|---|---|---|---|---|---|
| 🔴 | 17 | 🔵 | 16 | N | - | - | - |
| 🟢 | 17 | 🔴 | 17 | C | 289 | 15.83 | C5 |
| 🔵 | 16 | 🔴 | 17 | C | 272 | 14.50 | C6 |

## Results

Given our first goal was to develop a universal clustering algorithm that could recognize all kinds of clusters regardless of shape, size, or density, our priority was to test TC on as many different types of clustering problems as possible. Figure 3 presents the results of tests with nine different data sets containing seven challenges commonly faced in clustering. These data sets have been widely used as benchmark comparisons for many clustering algorithms [13]. As the results show, the proposed TC algorithm conquered every trial.

**Fig. 3. Results with seven different clustering challenges** As the results show, the proposed TC algorithm recognized all the clusters regardless of their shape, size, or density.

A  [14] **Highly overlapping data:** TC was easily able to recognize the 15 clusters in this data set with substantial overlaps.

B  [15] **A FLAME test case:** TC was able to find the two clusters in this case designed to test fuzzy clustering by local approximation of membership (FLAME).

C  [16] **Spectral path clustering:** This data set was used to illustrate the performance of a path-based spectral clustering algorithm. TC was perfectly able to identify the three clusters without the need to generate a connectivity graph.

D  [17] **Unbalanced data:** Severe imbalances in the data did not present a problem to TC as the hugely disproportionate clusters to the right show.

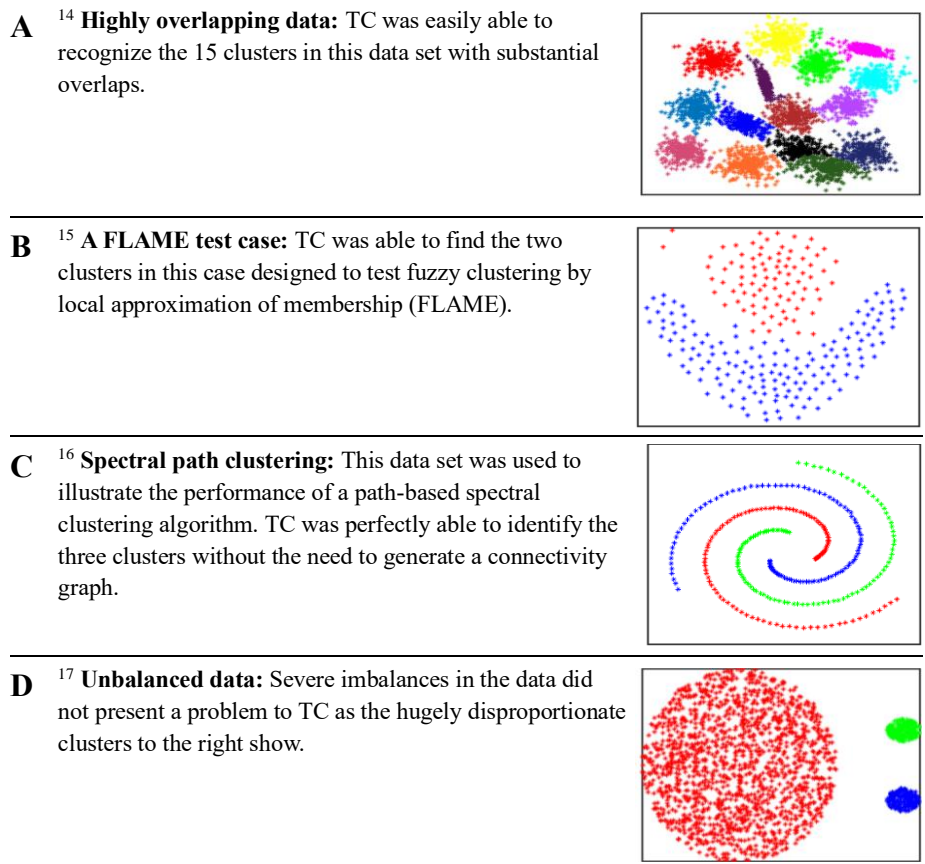

| | | |
|---|---|---|
| **E** | [18] **Noisy data:** This data was originally used to showcase how a density-based clustering algorithm (fast search and finding peaks in density) handles noise. TC was able to detect the five clusters. | 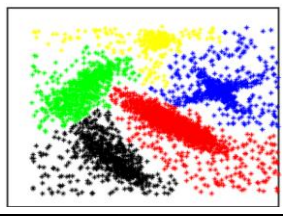 |
| **F** | [19] **Heterogeneous geometric properties:** TC intuitively found the three clusters without the need to calculate point symmetry distances as was required in [19]. | 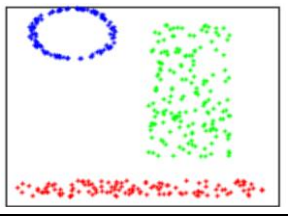 |

**Multi-objective clustering:** Figs. 3G-3I show examples of multi-objective clustering. With these types of tasks, more than one type of clustering algorithm is needed to reveal all the different types of cluster structures in the data [20]. The current standard is to use ensemble learning to optimize multiple objective functions. TC was able to identify the different structures naturally.

| **G** [21] | **H** [22] | **I** [23] |
|---|---|---|
| 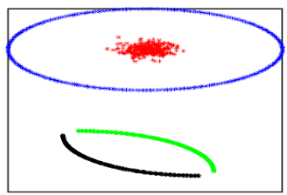 | 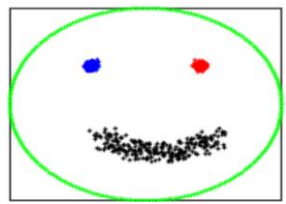 | 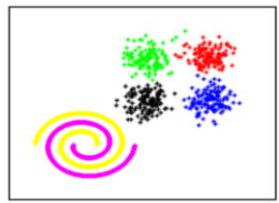 |

The corresponding decision graphs of these nine data sets are illustrated in Figs. S1A-I. Further, for comparison, we also ran these nine tests using K-means [24] and show the results in Fig S2. K-means failed on eight of the nine data sets, the exception being the first.

Moreover, we also compared the proposed TC algorithm with eight other representative clustering algorithms on these nine 2D data sets and evaluated the clustering quality using two external indices described in the supplementary materials. In these comparisons, the noisy dataset E was excluded because it contained no ground-truth labels, making it impossible to evaluate the clustering quality of the algorithms with external indices. The comparison results (Tables S2-S5) show that TC achieved the highest accuracy on six of the eight data sets, the second-highest accuracy on a seventh, and 3rd and 4th in terms of two different indices on the eighth. Notably, the algorithms that outperformed TC on the 7th and 8th data sets are sensitive to initialization/parameters, so the results reported in Table S5 are the highest accuracy from many runs with different initializations/parameter settings. TC, however, is parameter-free and does not need any initialization, so its reported accuracy levels are from a single run.

In our second set of experiments, we tested TC with three different clustering applications: gene expression, face recognition and handwritten-digits recognition.

To assess TC's efficacy for gene expression analysis [25], we chose the RNA-Seq (HiSeq) PANCAN data set from [26], which is a random extraction of gene expressions from patients with different types of tumors: BRCA, KIRC, COAD, LUAD and PRAD. The data set contains 801 samples and 20,531 dimensions (genes). TC was able to classify the patients according to the five different types of tumors with 99.88% accuracy.

The decision graphs are shown in Fig. S3, which reveal four abnormal connections. Once these were removed as part of the standard procedure, the appropriate number of five clusters remained. A more intuitive representation appears in Fig. 4A with a plot of the 20,531-dimensional feature space in 3D space using principal component analysis (PCA) [27]. We have also provided five gene heatmaps corresponding to the five types of tumors, as shown in Figs. 4B-F.

**Fig. 4. Gene heatmaps** of the five types of tumors in the RNA-Seq (HiSeq) PANCAN data set [26] clustered by TC. To better represent TC's clustering power, we calculated the mutual information values between one of the 20,531 genes in the original data set and the final cluster assignment labels. The 30 genes with the highest mutual information values are shown here. In each heatmap, the horizontal axis represents the genes, and the vertical axis represents the samples. Different color depths represent the different gene expression values.

**A**

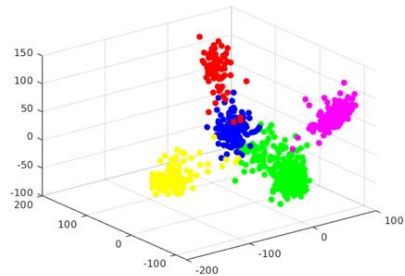

Projection of the five clusters (tumors) found by TC in a three-dimensional subspace.

Figs 4B-4E show the gene heatmaps of the five types of tumors. The initials next to the figure letter are the name of the tumor, followed by the number of samples.

**B** PRAD: 136 samples

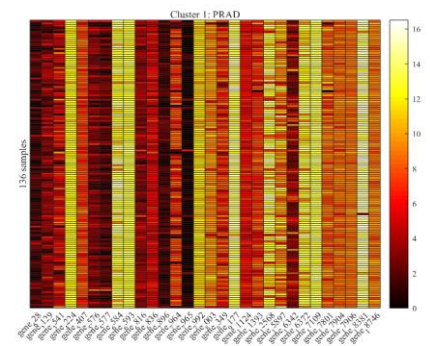

**C** LUAD: 140 samples

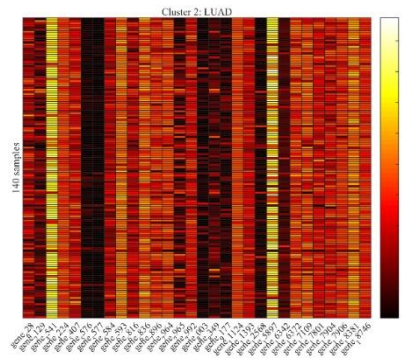

**D** BRCA: 301 samples

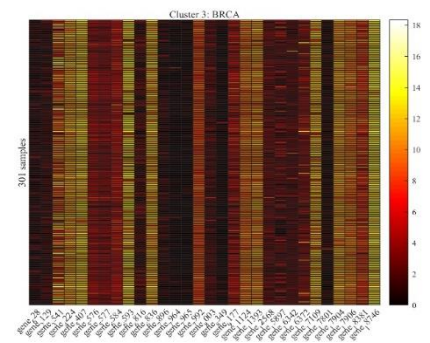

**E** KIRC: 146 samples

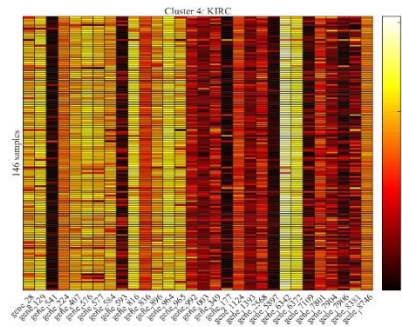

**F** COAD: 78 samples

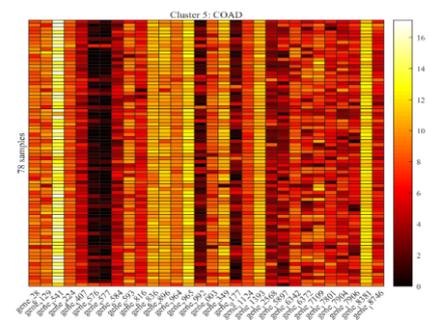

Our chosen face recognition data set was the Olivetti Face Database [28]. Here, we compared TC to the DPC algorithm. DPC is an unsupervised clustering method based on fast search and finding peak densities [18]. With both algorithms, we measured the similarity between two images following the method in ref. [29].

Using the first 100 images in the data set only for ease of reporting, TC completed the task with 95% accuracy, while DPC finished with 87%. The images color-coded by cluster appear in Fig. S4, and the corresponding decision graph appears in Fig. S5A. DPC has a hyper-parameter $d_c$, which is hard to be known in advance and needs to be adjusted many times, but TC is parameter-free. Yet even when we used the best setting ($d_c = 0.07$) recommended in the original article [18] and manually determined the ideal number of clusters to be 10 (based on the ground-truth, but incorrectly identified as nine in the article), we could not exceed 87%. Comparing TC's decision graph to DPC's in Fig. S5B, the ease of estimating the ideal number of clusters is clear.

For the last task, the handwritten digit recognition, we chose the popular benchmark MNIST [30] with 10,000 samples and 4,096 dimensions. Each digit should be clustered, making 10 clusters in total from 0 to 9. Our comparator was FINCH (Efficient Parameter-free Clustering Using First Neighbor Relations) [11], and our evaluation metric was normalized mutual information (NMI) [31]. Correctly classifying 97.67% of the digits, TC was more accurate than FINCH and the other 11 clustering algorithms reported in ref. [11]. Moreover, the nine abnormal connections that needed to be removed to leave the correct 10 clusters are clearly visible in TC's decision graph (see Fig. S6). FINCH, however, requires users to subjectively choose the "ideal" partitions from several options. In this case, five partitions were offered (Partition 1: 1699 clusters, Partition 2: 310 clusters, Partition 3: 65 clusters, Partition 4: 17 clusters, Partition 5: 10 clusters), not the 10 clusters required by the task.

Similarly, we also compared TC with eight other representative clustering algorithms on the above three tasks, and TC still achieved the highest accuracy of all of them (Tables S2-S5).

**Discussion**

TC requires users to determine and remove the abnormal connections from the decision graph. As the above experiments show, in many scenarios, identifying outlying clusters partitions is easy. Alternatively, sometimes the ground-truth number of clusters K is known. Therefore, K-1 connections with the largest $M_i \times D_i$ can simply be removed to arrive at the correct number of clusters. A third strategy is to analyze the rough cluster partitions in each level of the hierarchical tree, as shown in Table S6, and make decisions from there.

However, there may still be some very difficult cases where all three of these strategies do not work well. The gene expression analysis shown in Figure 4 is one such case. Unlike the other test cases above, issues with data sparsity meant it was not immediately clear from the decision graph (Fig. S3) as to whether there were four or five abnormal connections. In such situations, clues for accurately determining which connections are truly abnormal can be found by plotting $\gamma = M_i \times D_i$ sorted in decreasing order (Fig. S7). The graph shows that this quantity starts growing anomalously below a rank order of 4, which suggests the first four connections should be removed to leave five clusters.

Another difficult case is severely imbalanced data where some clusters contain very few points or the distances between some clusters and their 1-nearest neighbor is too small. In the decision graph, the corresponding connections will appear very close to each other on the coordinate axis, making it difficult to identify them clearly. To address this, we propose a simple scheme for automatically determining abnormal connections. Eq. (1) will reveal many connections $C_i$ throughout the entire clustering process and, as we know, each $C_i$ has two properties, $M_i$ and $D_i$. Therefore, abnormal connections (ANC) could be defined as

$$ANC = \{C_i | M_i \geq meanM \cap D_i \geq meanD \} \quad (4)$$

where *meanM* is the mean value of all $M_i$, and *meanD* is the mean value of all $D_i$.

To test whether this scheme works, we benchmarked TC on two classic data sets: the Columbia University Image Library (COIL-100) [32] and Shuttle. COIL-100 is a classic collection of color pictures of 100 objects, each imaged from 72 viewpoints for a total of 7,200 samples and 49,152 dimensions. Shuttle is a data set from NASA that contains 58,000 multivariate measurements produced by the sensors in the radiator subsystem of a space shuttle. The measurements are known to be caused by seven different conditions of the radiators [1].

Further, in addition to TC, we also tested one of the most recent clustering algorithms, RCC (Robust Continuous Clustering) [1], for comparison. The metric used was adjusted mutual information (AMI) [33]. As the decision graphs for the two data sets in Figs. S8A and S8B show, many connections are very close to the coordinate axis, which makes them difficult to identify. However, applying Eq. (4) to automatically identify and remove the abnormal connections left 97 clusters with an AMI of 97.21% for the COIL-100 data set and 6 clusters with an AMI of 60.75% for the Shuttle data set – very close to the ground-truths of 100 and 7, respectively. The AMIs are not ideal, but they are still better than RCC and the other 14 state-of-the-art clustering algorithms reported in ref. [1]. In addition, for completeness, we also compared TC with eight other representative clustering algorithms on the above two data sets. Similarly, TC achieved the highest accuracy among them all (Tables S2-S5).

Whether compared with classic or very recent methods, TC demonstrates itself to be a highly accurate algorithm, superior in performance to all its counterparts and with an unprecedented level of universality. In summary, we have presented a clustering algorithm that is: parameter-free, can recognize all kinds of clusters regardless of their shapes, size or densities; does not depend on a priori knowledge; is robust to noise and outliers; does not need any initialization; and does not demand a manually-specified stopping condition. Moreover, the ability to use any desired method of 1-nearest-distance computation means TC scalable to large data sets with low computational overhead and reasonable time complexity, especially when choosing fast approximate nearest neighbor methods, such as k-d tree or locality-sensitive hashing [11,12]. Above, we presented many test cases to showcase TC's versatility, and even more comparisons of clustering quality with the state-of-the-art methods are provided in Tables S2-S5 in the supplementary materials.

**Acknowledgments**

This work was supported by the Australian Research Council (ARC) under discovery grant DP180100670 and DP180100656, the Australia Defence Innovation Hub under Contract No. P18-650825 and the US Office of Naval Research Global under Cooperative Agreement Number ONRG - NICOP - N62909-19-1-2058. We would also like to thank the NSW Defence Innovation Network and the NSW State Government for financial support of part of this research through grant DINPP2019 S1-03/09.

**Supplementary Materials**

    Materials and Methods

    Supplementary Text

    Figs. S1 to S8

    Tables S1 to S6

# Supplementary Materials for

## Clustering via torque balance with mass and distance

Jie Yang and Chin-Teng Lin

University of Technology Sydney, Faculty of Engineering and Information Technology,
Centre for Artificial Intelligence, 15 Broadway, Ultimo, New South Wales 2007, Australia

Email: jie.yang-4@student.uts.edu.au; Chin-Teng.Lin@uts.edu.au

**This PDF file includes:**

Materials and Methods
Supplementary Text
Figs. S1 to S8
Tables S1 to S6

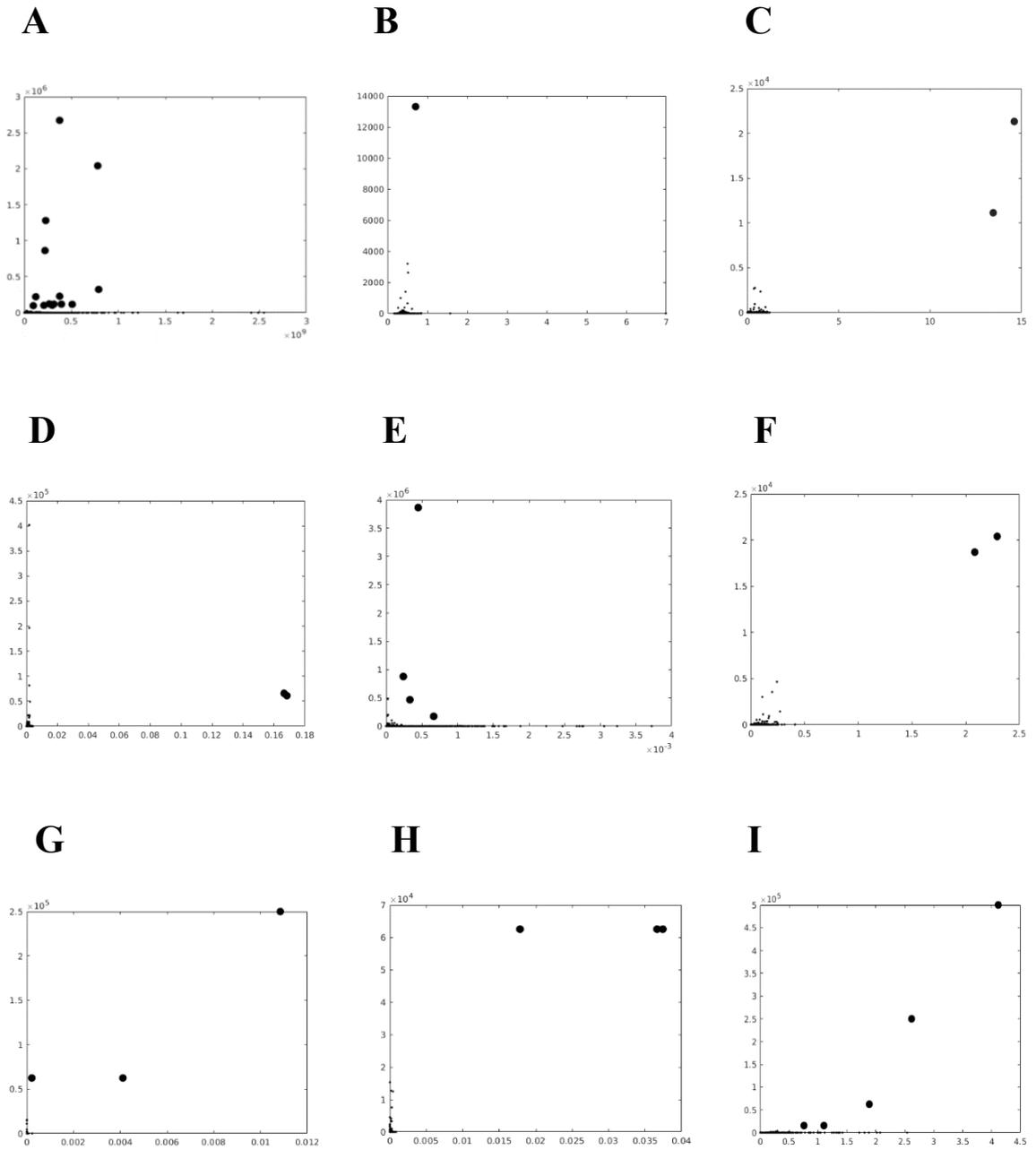

**Fig. S1. TC decision graphs for the data point distributions in Fig. 3** with $d^2$ on the horizontal axis and $m_1 \times m_2$ on the vertical axis. Abnormal connections appear in bold. Removing these connections leaves the final cluster partitions for each data set.

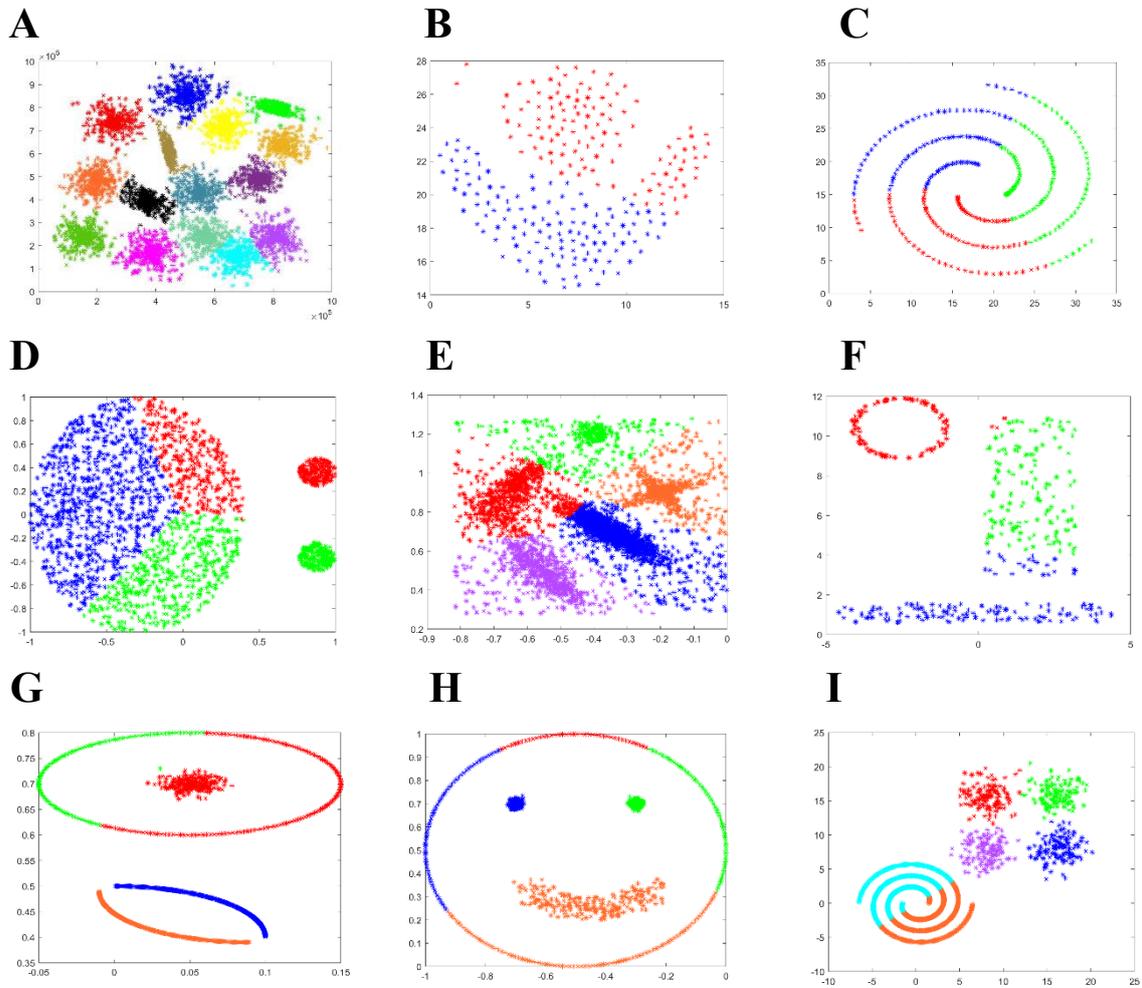

**Fig. S2. K-means comparisons to the nine experiments in Fig. 3.** The results reported are the best solution from 100 runs according to the ground-truth labels or, in the case of data set E, the objective function as it does not contain ground-truth labels. The initialization method was the well-known K-means++ method [34], and the value of K was set to the ground-truth number of clusters.

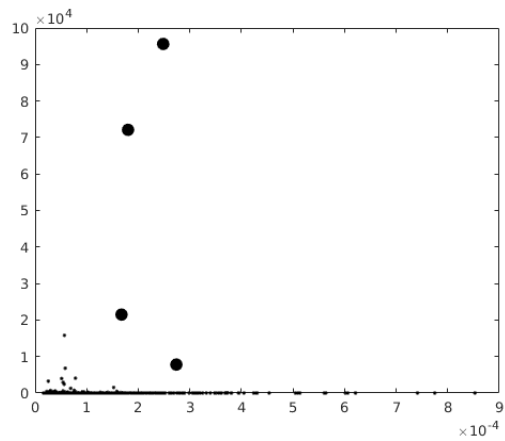

**Fig. S3. TC decision graph on the RNA-Seq (HiSeq) PANCAN data set** [26] with $d^2$ on the horizontal axis and $m_1 \times m_2$ on the vertical axis. Removing the four abnormal connections in bold leaves five clusters corresponding to the five types of tumors.

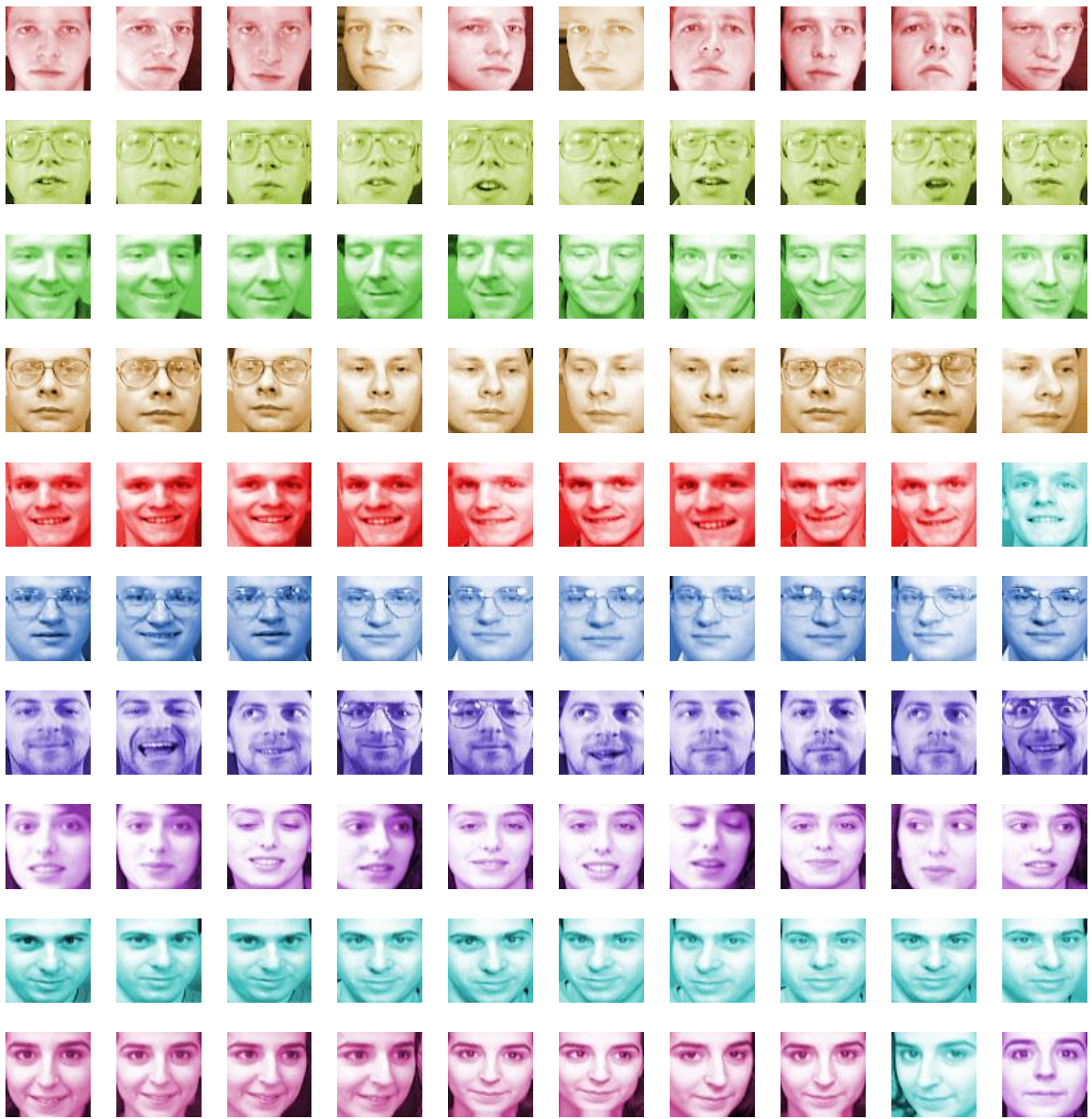

**Fig. S4 Cluster analysis of the first 100 images of the Olivetti Face Database** [28]. Faces with the same color wash belong to the same cluster. TC's accuracy was 95%, higher than the 87% delivered by the DPC method [18].

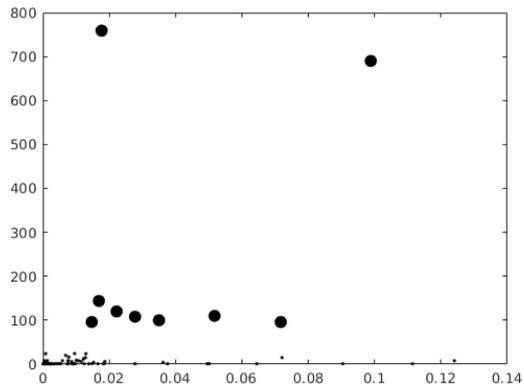 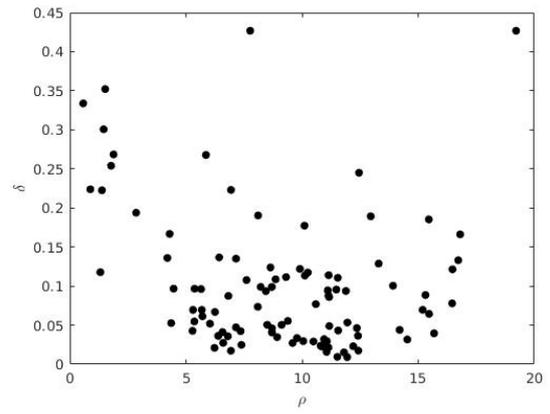

**Fig. S5. Decision graphs for the Olivetti Face Database tests.** **(A)** shows TC; **(B)** shows DPC [18], with $d^2$ on the horizontal axis and $m_1 \times m_2$ on the vertical axis in **(A)**, and with data density on the horizontal axis and density-relative distance [18] on the vertical axis in **(B)**. In **(A)**, the nine abnormal connections appear in bold, and removing them leaves the 10 ground-truth clusters. However, identifying the correct 10 clusters from the DPC decision graph in **(B)** would be extremely difficult, if not impossible.

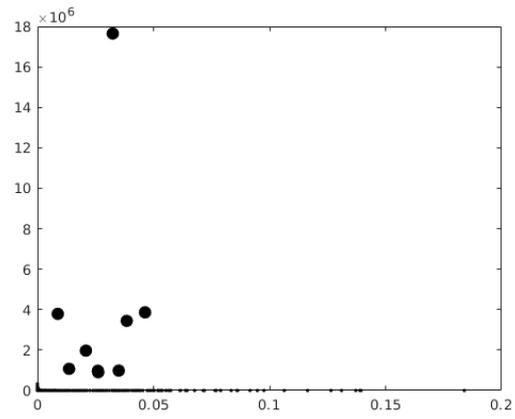

**Fig. S6. TC decision graph for the MNIST handwritten digits data set** [30] with $d^2$ on the horizontal axis and $m_1 \times m_2$ on the vertical axis. Nine abnormal connections appear in bold, two of which have some overlap. Removing them leaves correct number of 10 clusters.

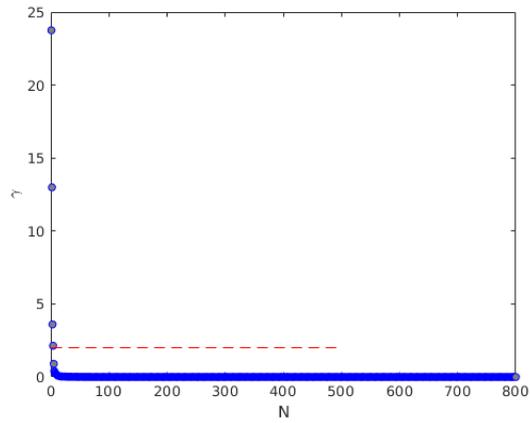

**Fig. S7. A method for distinguishing ambiguous abnormal connections.** This decision graph plots $\gamma = (m_1 \times m_2) \times d^2$ for each connection in the RNA-Seq (HiSeq) PANCAN data set in descending order. The graph shows that this quantity starts growing anomalously below a rank order of 4, which suggests the first four connections should be removed to leave five tumor clusters.

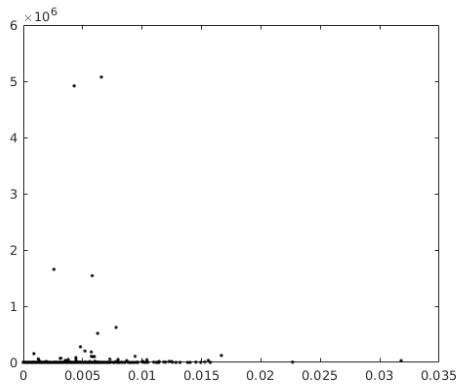 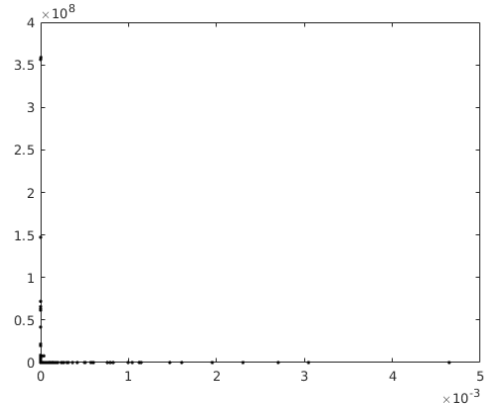

**Fig. S8. Difficulties identifying abnormal connections with sparse data.** **(A)** shows the TC decision graph for the COIL-100 data set [32] and **(B)** for the Shuttle data set (both with $d^2$ on the horizontal axis and $m_1 \times m_2$ on the vertical axis). When clusters at the end of the connections have too few points, the small values of $m_1 \times m_2$ will place the connections too close to the x-axis to easily discern **(A)**. Similarly, small distances between the clusters and their 1-nearest cluster will plot the $d^2$ of the connections too close to the y-axis **(B)**. These will both result in the corresponding connections being very close to the coordinate axis in the decision graphs. In these situations, Eq. (4) can be used to automatically determine the abnormal connections that need to be removed.

**Table S1. The pseudocode of the proposed torque clustering (TC) algorithm**

**The Torque-Clustering Algorithm**:

1: **Input**: Distance matrix $S \in R^{n \times n}$, where $S$ is a symmetric matrix containing the distance between any pair of samples and $n$ denotes the total number of samples. The data set $X \in R^{n \times d}$, where $n$ denotes the total number of samples and each sample point is represented by $d$ attributes or feature dimensions.

2: **Output**: Cluster partition $\phi = \{\zeta_1, \zeta_2, \ldots, \zeta_K\}$.

3: Regard each sample as a cluster to get the initial cluster partition $\{\zeta_1, \zeta_2, \ldots, \zeta_n\}$.

4: Calculate the mass value of all clusters $\{mass_1, mass_2, \ldots, mass_n\}$, which equals the number of samples each cluster contains. Currently, each $mass_i$ equals to 1.

5: Calculate the 1-nearest clusters $\{\zeta_1^N, \zeta_2^N, \ldots, \zeta_n^N\}$ of all $\zeta_i$ and corresponding distances $\{d(\zeta_1, \zeta_1^N), d(\zeta_2, \zeta_2^N), \ldots, d(\zeta_n, \zeta_n^N)\}$ between $\zeta_i$ and its 1-nearest cluster $\zeta_i^N$ according to the distance matrix $S$ or using fast approximate nearest neighbor methods (such as kd-tree or locality-sensitive hashing).

6: Construct the connected graph G according to the rule in Eq. (1) and calculate the connected component of the connected graph to obtain the new cluster partition $\phi = \{\zeta_1, \zeta_2, \ldots, \zeta_L\}$, where $L < n$.

7: Calculate the two properties of each connection according to Eq. (2) and Eq. (3), and save these to M and D, respectively, where M contains the property values $mass(\zeta_i) \times mass(\zeta_i^N)$, and D contains the property values $d^2(\zeta_i, \zeta_i^N)$.

8: **While** there are at least two clusters in new cluster partition $\phi$ **do**

9: Calculate the mass value of all cluster $\{mass_1, mass_2, \ldots, mass_L\}$.

10: Calculate the 1-nearest clusters $\{\zeta_1^N, \zeta_2^N, \ldots, \zeta_L^N\}$ of all $\zeta_i$ and corresponding distances $\{d(\zeta_1, \zeta_1^N), d(\zeta_2, \zeta_2^N), \ldots, d(\zeta_n, \zeta_L^N)\}$ between $\zeta_i$ and its 1-nearest cluster $\zeta_i^N$ using the single-linkage method or other linkage methods according to the distance matrix $S$. Or, compute the mean (average of all data vectors in that cluster) for each cluster in new cluster partition and use the mean to represent each cluster, then calculate the 1-nearest clusters $\{\zeta_1^N, \zeta_2^N, \ldots, \zeta_L^N\}$ of all $\zeta_i$ and corresponding distances $\{d(\zeta_1, \zeta_1^N), d(\zeta_2, \zeta_2^N), \ldots, d(\zeta_n, \zeta_L^N)\}$ between $\zeta_i$ and its 1-nearest cluster $\zeta_i^N$ using fast approximate nearest neighbor methods.

11: Update the connected graph G according to the rule in Eq. (1) and calculate the connected component of the connected graph to obtain the new cluster partition $\phi$.

12: Calculate the two properties of each connection according to Eq. (2) and Eq. (3), and save these to M and D, respectively.

13: **end While**

14: Observe the decision graph plotted by M and D or use Eq. (4) to determine the connections that need to be removed.

15: Calculate the connected component after removing the connections of connected graph G determined in Step 14 to obtain the final cluster partition $\phi = \{\zeta_1, \zeta_2, \ldots, \zeta_K\}$.

# Comparison between TC and 8 representative clustering algorithms

We benchmarked TC on 13 of the 14 data sets used throughout this study against 8 representative clustering algorithms.

The descriptive statistics of the data sets are given in Table S2. We did not include the 2D_Data5 because it has no ground-truth labels, which makes it impossible to evaluate the clustering quality of algorithms with external indices. We calculated similarity in the 2D data sets according to Euclidean distance as is common practice and as cosine distance for other data sets, with the exception of the Olivetti face database (OFD-F100), where we followed the method outlined in ref. [29].

**Table S2. Statistics of the 13 data sets**

Imbalance is defined as the ratio of the largest and smallest cardinalities of the ground-truth clusters. OFD-F100 stands for the first 100 images of the Olivetti Face Database. RNA-Seq means the RNA-Seq (HiSeq) PANCAN. The data set 2D_Data5 was removed because it contained no ground-truth labels, making it impossible to evaluate the clustering quality of the algorithms with external indices.

| Data set | Instances | Dimensions | No. of ground-truth clusters | Imbalance |
|---|---|---|---|---|
| 2D_Data1 | 5000 | 2 | 15 | 1.1672 |
| 2D_Data2 | 240 | 2 | 2 | 1.7586 |
| 2D_Data3 | 312 | 2 | 3 | 1.0495 |
| 2D_Data4 | 2000 | 2 | 3 | 8 |
| 2D_Data6 | 400 | 2 | 3 | 1.5455 |
| 2D_Data7 | 1000 | 2 | 4 | 1 |
| 2D_Data8 | 1000 | 2 | 4 | 1 |
| 2D_Data9 | 1500 | 2 | 6 | 4.0650 |
| OFD-F100 | 100 | 64×64 | 10 | 1 |
| MNIST | 10000 | 4096 | 10 | 1.2724 |
| COIL-100 | 7200 | 49152 | 100 | 1 |
| Shuttle | 58000 | 9 | 7 | 4558 |
| RNA-Seq | 801 | 20531 | 5 | 3.8462 |

Details of the eight baselines chosen for comparison follow. The parameter settings for those algorithms that require them are given in Table S3.

**K-means++** [34]

We used the implementation provided in Matlab and, as recommended, report the best index values at termination from 100 runs.

**Gaussian mixture model clustering (GMM)**

We used the implementation provided by ref. [35] and, again, report the best index values at termination from 100 runs as recommended.

**Fuzzy clustering (Fuzzy)** [36]

We used the Matlab implementation and report the average results across 100 random restarts.

**Spectral clustering (SC)** [37, 38]

As with Fuzzy, we used the Matlab implementation and report the average results across 100 random restarts.

**Hierarchical clustering average-linkage (HAC-Avg)**
The implementation came from Matlab.
**Hierarchical clustering ward-linkage (HAC-Ward)** [7]
As above, the implementation came from Matlab.

We also chose two very recent algorithms.
**Clustering by fast search and find of density peaks (DPC)** [18]
We followed the implementation recommended in the original paper [18]. DPC has a parameter, *dc*, which is used to calculate the density of samples. To maximize the clustering quality, we followed the guidelines in the paper and tested $dc = 1.0\%, 1.1\%, 1.2\%, 1.3\%, 1.4\%, 1.5\%, 1.6\%, 1.7\%, 1.8\%, 1.9\%, 2\%$, then choose the best setting for each data set. Further, in some data sets, it is very difficult to determine the optimal number of clusters from the decision graphs alone (see Fig. S5B). Hence, to maximize DPC's clustering quality with these data sets, we used the ground-truth number of clusters instead of referring to the decision graphs.

**Efficient parameter-free clustering using first neighbor relations (FINCH)** [11] With this algorithm, we also followed the implementation recommended in the original paper [11]. FINCH is a parameter-free algorithm that provides several options cluster partitions and asks the user to make a subjective decision as to which is the "ideal" scheme. As an example, FINCH generated five partitioning schemes with the MNIST image set ranging from 1,699 clusters right down to 10 clusters. Therefore, to maximize clustering quality, we chose the scheme closest to the ground-truth number of clusters, which accords with the authors' approach [11]. FINCH's ability to estimate the optimal number of clusters compared to TC is shown in Table S4.

**Table S3. Parameter settings**

| Algorithm | Parameter settings |
|---|---|
| K-means++ | EmptyAction= 'singleton', MaxIter= 100, Replicates= 1, K= ground-truth number of clusters for each data set |
| GMM | tol = $10^{-10}$, maxiter = 500, required number of clusters = ground-truth number of clusters for each data set |
| Fuzzy | exponent for the matrix U= 2.0, MaxIter= 100, threshold= $10^{-5}$, required number of clusters = ground-truth number of clusters for each data set |
| SC | LaplacianNormalization= 'randomwalk', SimilarityGraph= 'knn', NumNeighbors= 10, KNNGraphType= 'complete', ClusterMethod= 'kmeans' |
| HAC-Avg | 'cutoff'= ground-truth number of clusters for each data set |
| HAC-Ward | 'cutoff'= ground-truth number of clusters for each data set |
| DPC | $dc \in \{1.0\%, 1.1\%, 1.2\%, 1.3\%, 1.4\%, 1.5\%, 1.6\%, 1.7\%, 1.8\%, 1.9\%, 2\%\}$, ideal number of clusters= ground-truth number of clusters for each data set |

**Table S4. Ability to estimate the optimal number of clusters.** A comparison between TC and FINCH shows TC accurately estimated the optimal number of clusters on 10 of 12 data sets, while FINCH accurately estimated only one. #C is the ground-truth number of clusters for each data set. Correct estimations are highlighted in bold.

| | 2D_Data1 | 2D_Data2 | 2D_Data3 | 2D_Data4 | 2D_Data6 | 2D_Data7 | 2D_Data8 | 2D_Data9 | MNIST | COIL-100 | Shuttle | RNA-Seq |
|---|---|---|---|---|---|---|---|---|---|---|---|---|
| #C | 15 | 2 | 3 | 3 | 3 | 4 | 4 | 6 | 10 | 100 | 7 | 5 |
| TC_estimate | **15** | **2** | **3** | **3** | 3 | **4** | **4** | **6** | **10** | 97 | 6 | **5** |
| FINCH_estimate | 56 | 19 | 56 | 128 | 36 | 26 | 63 | 7 | **10** | 44 | 16 | 4 |

All experiments were performed in Matlab2019b, and performance was evaluated against the two well-known metrics: NMI [31, 39] and ACC [40]. The results for all baselines on all data sets are reported in Tables S5A and B, with the maximum values highlighted in bold. We also ranked each algorithm according to its average performance across all data sets. For example, if an algorithm achieves the third-highest accuracy on half of the data sets and the fourth-highest one on the other half, its average rank would be 3.5 (3 × 0.5 + 4 × 0.5). If an algorithm did not scale to a data set, that data set was not taken into account in the ranking calculation.

**Table S5. Comparison performance of all algorithms of 13 data sets.** "NA" means not applicable and indicates that the algorithm did not scale to the data set. For example, most algorithms cannot be applied to the data set OFD-F100, because it only provides the distance matrix and not the coordinates of each sample. GMM and HAC-Ward were not applied to the data sets COIL-100 and Shuttle because these two algorithms are not scalable to the data sets with large data size.

**A**

**Results in terms of NMI.**

| Data set | K-means++ | GMM | Fuzzy | SC | HAC-Avg | HAC-Ward | DPC | FINCH | TC |
|---|---|---|---|---|---|---|---|---|---|
| 2D_Data1 | .9652 | .9508 | .9580 | .0128 | .9498 | .9480 | **.9747** | .8665 | .9568 |
| 2D_Data2 | .4843 | .4477 | .4422 | .0479 | .4044 | .3905 | .4132 | .4896 | **1** |
| 2D_Data3 | .0012 | .0678 | .0003 | 1 | .0294 | .0017 | 1 | .5359 | **1** |
| 2D_Data4 | .4453 | 1 | .4429 | 1 | .4061 | .6183 | 1 | .3720 | **1** |
| 2D_Data6 | .8089 | 1 | .8016 | 1 | .8529 | .7110 | .7445 | .5583 | **1** |
| 2D_Data7 | .8357 | .8406 | .6008 | 1 | .7201 | .6048 | .8044 | .6995 | **1** |
| 2D_Data8 | .6807 | .9448 | .6072 | 1 | .6955 | .6205 | .6663 | .6846 | **1** |
| 2D_Data9 | .7065 | .7109 | .5319 | .7881 | .7020 | .7229 | **.9950** | .7947 | .9925 |
| OFD-F100 | NA | NA | NA | .8132 | NA | NA | .8666 | NA | **.9343** |
| MNIST | .9741 | .8396 | .3338 | .9761 | .9370 | .9263 | .9751 | .9755 | **.9767** |
| COIL-100 | .8281 | NA | .2866 | .8564 | .7526 | .8381 | .8657 | .7897 | **.9820** |
| Shuttle | .3247 | .3888 | .2415 | .5860 | .3023 | NA | .5769 | .0368 | **.6075** |
| RNA-Seq | .9808 | .8400 | .5711 | .9948 | .8696 | .8477 | .8348 | .8785 | **.9948** |
| *Rank* | 4.9 | 4.3 | 7.5 | 3.0 | 6.0 | 6.9 | 4.1 | 5.4 | **1.4** |

**B**

**Results in terms of ACC.**

| Data set | K-means++ | GMM | Fuzzy | SC | HAC-Avg | HAC-Ward | DPC | FINCH | TC |
|---|---|---|---|---|---|---|---|---|---|
| 2D_Data1 | .9808 | .9128 | .9653 | .0773 | .9680 | .9654 | **.9858** | .7468 | .9714 |
| 2D_Data2 | .8583 | .8417 | .8500 | .6458 | .8000 | .7708 | .7875 | .2292 | **1** |
| 2D_Data3 | .3526 | .4295 | .3401 | 1 | .4167 | .3526 | 1 | .1571 | **1** |
| 2D_Data4 | .6150 | 1 | .4732 | 1 | .4955 | .6085 | 1 | .1015 | **1** |
| 2D_Data6 | .9325 | 1 | .9325 | 1 | .9450 | .8100 | .8500 | .1350 | **1** |
| 2D_Data7 | .8530 | .8870 | .6055 | 1 | .6120 | .6820 | .6560 | .4280 | **1** |
| 2D_Data8 | .7910 | .9820 | .6860 | 1 | .8140 | .7460 | .7320 | .5490 | **1** |
| 2D_Data9 | .7280 | .7607 | .5179 | .5072 | .7153 | .7840 | **.9987** | .6133 | .9980 |
| OFD-F100 | NA | NA | NA | .7447 | NA | NA | .7800 | NA | **.9500** |
| MNIST | .9915 | .7842 | .2086 | .9921 | .9678 | .9615 | .9916 | .9918 | **.9922** |
| COIL-100 | .5992 | NA | .0233 | .6368 | .3344 | .6158 | .5482 | .3922 | **.9383** |
| Shuttle | .5456 | .5116 | .3000 | .7581 | .4248 | NA | .8893 | .5788 | **.8941** |
| RNA-Seq | .9950 | .8939 | .5546 | .9988 | .9164 | .9151 | .7990 | .8240 | **.9988** |
| *Rank* | 4.2 | 4.1 | 7.2 | 3.4 | 5.2 | 5.8 | 4.2 | 7.9 | **1.3** |

The results overwhelmingly support TC as a universal solution to clustering given its best-in-show performance on 11 of the data sets and second place on the 2D_Data9 data set. On the 2D_Data1 data set, TC ranked 3rd and 4th in terms of the ACC and NMI indices, respectively. The algorithms that outperformed TC on these two data sets (K-means++, Fuzzy, DPC) are sensitive to initialization/parameters, so the reported accuracies are the highest of 100 runs with different initializations for the K-means++ and Fuzzy algorithms and 11 runs with different parameters for the DPC algorithm. TC, however, is parameter-free and does not need any initialization, so the accuracy levels reported are from just a single run. In terms of rank, the next-best algorithm, SC, was more than double TC. Moreover, for the data sets 2D_Data1, 2D_Data2, 2D_Data9, OFD-F100, and COIL-100, the clustering quality of SC is much worse than TC's. Combined, these two results demonstrate TC to be a clustering algorithm with extremely high accuracy and versatility.

**Table S6. Hierarchical trees for each data set.**

Table S6 shows the number of clusters in each layer of the hierarchical tree for each data set produced by the proposed TC algorithm. In Step 0, the number of clusters equals the number of samples but, by the end of the process, each data set will have been merged into one giant cluster. The number of clusters in each step can also provide useful information for determining abnormal connections with ambiguous decision graphs when *a priori* knowledge of the rough number of clusters in the target data set is available. For example, if we know that the proper number of clusters for 2D_Data1 is around 15-20, then the 19 clusters generated in Step 6 would be a good starting point. Examining this decision graph may reveal an abnormal connection that, when removed, would leave the proper number of clusters. Notably, TC's hierarchies average 8.4 mergers (layers) across the 14 data sets used in this study. A conventional hierarchical clustering algorithm would require an average of 6,538.5 mergers because it needs to merge each sample n-1 times, where n is the number of samples in the data set.

| | Step0 | Step1 | Step2 | Step3 | Step4 | Step5 | Step6 | Step7 | Step8 | Step9 | Step10 | Step11 | Step12 |
|---|---|---|---|---|---|---|---|---|---|---|---|---|---|
| 2D_Data1 | 5000 | 1530 | 618 | 267 | 99 | 38 | 19 | 9 | 5 | 4 | 1 | - | - |
| 2D_Data2 | 240 | 63 | 22 | 9 | 4 | 2 | 1 | - | - | - | - | - | - |
| 2D_Data3 | 312 | 113 | 43 | 21 | 10 | 6 | 4 | 2 | 1 | - | - | - | - |
| 2D_Data4 | 2000 | 626 | 259 | 118 | 54 | 25 | 10 | 4 | 1 | - | - | - | - |
| 2D_Data5 | 4000 | 1217 | 490 | 209 | 98 | 47 | 22 | 8 | 4 | 2 | 1 | - | - |
| 2D_Data6 | 400 | 126 | 51 | 28 | 14 | 9 | 5 | 3 | 1 | - | - | - | - |
| 2D_Data7 | 1000 | 342 | 146 | 65 | 32 | 14 | 6 | 4 | 2 | 1 | - | - | - |
| 2D_Data8 | 1000 | 332 | 153 | 69 | 33 | 16 | 12 | 9 | 6 | 4 | 1 | - | - |
| 2D_Data9 | 1500 | 164 | 70 | 31 | 13 | 5 | 3 | 2 | 1 | - | - | - | - |
| OFD-100 | 100 | 31 | 11 | 4 | 1 | - | - | - | - | - | - | - | - |
| MNIST | 10000 | 1699 | 528 | 155 | 54 | 19 | 9 | 4 | 3 | 1 | - | - | - |
| COIL-100 | 7200 | 2211 | 975 | 473 | 241 | 119 | 53 | 18 | 8 | 2 | 1 | - | - |
| Shuttle | 58000 | 15884 | 5963 | 2362 | 928 | 375 | 163 | 78 | 36 | 12 | 5 | 2 | 1 |
| RNA-Seq | 801 | 80 | 18 | 7 | 3 | 2 | 1 | - | - | - | - | - | - |